%% file: main.tex
\documentclass[10pt,twocolumn,letterpaper]{article}
\usepackage[pagenumbers]{cvpr}
\input{preamble}

\title{\mbox{World-$\delta$}: Continually Integrating Observations for \\ Incremental 3D Scene Generation}
\author{%
  Haonan Wang$^{1,2}$ \quad Hanyu Zhou$^{3}$\thanks{Corresponding author} \quad Xiaobin Hu$^{3}$ \quad Luxin Yan$^{1}$ \\
  $^{1}$School of Artificial Intelligence and Automation, Huazhong University of Science and Technology \\ 
  $^{2}$Forma AI, Xingtai Qiyuan Co., Ltd. \\ 
  $^{3}$School of Computing, National University of Singapore \\[0.3em]
  {\tt\small \{whn\_aurora,yanluxin\}@hust.edu.cn, hy.zhou@nus.edu.sg}
}
\hypersetup{pdftitle={World-delta: Continually Integrating Observations for Incremental 3D Scene Generation},pdfauthor={Haonan Wang, Hanyu Zhou, Xiaobin Hu, Luxin Yan}}
\begin{document}
\maketitle

\suppressfloats[t]

\SchedulePageFloats{1}{%
\begin{figure}[t]
  \centering
  \setlength{\abovecaptionskip}{0.1cm}
  \setlength{\belowcaptionskip}{-0.5cm}
  \includegraphics[width=\linewidth]{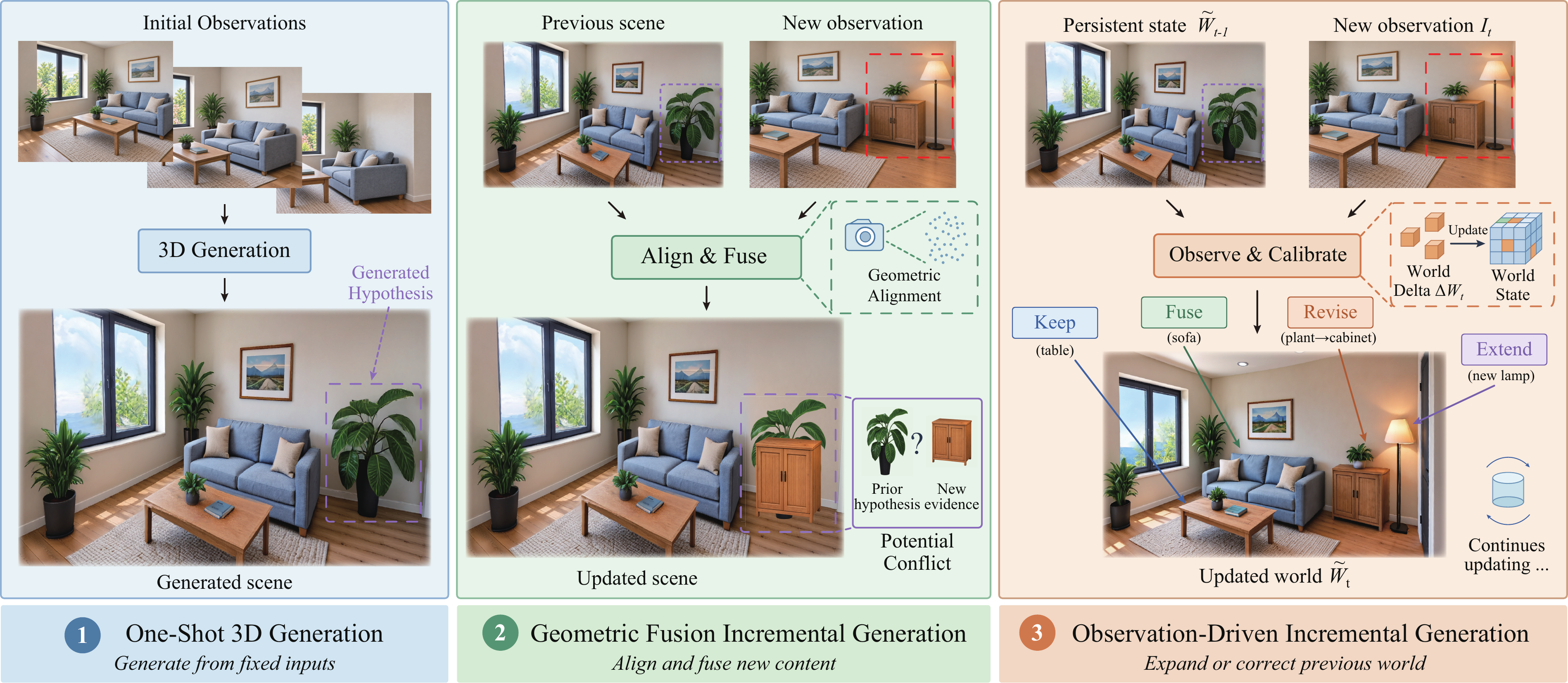}
  \caption{\textbf{Observation-driven 3D generation.}
  One-shot methods complete scenes from fixed inputs, whereas geometric fusion
  can retain conflicts with earlier completions. World-$\delta$ uses new
  observations to expand scene coverage and revise provisional hypotheses.
  A persistent world state and selective deltas preserve supported content
  while integrating new evidence.}
  \label{fig1}
\end{figure}%
\begin{figure*}[!t]
	\centering
	\setlength{\abovecaptionskip}{0.3cm}
	\setlength{\belowcaptionskip}{-0.1cm}
	\includegraphics[width=\linewidth]{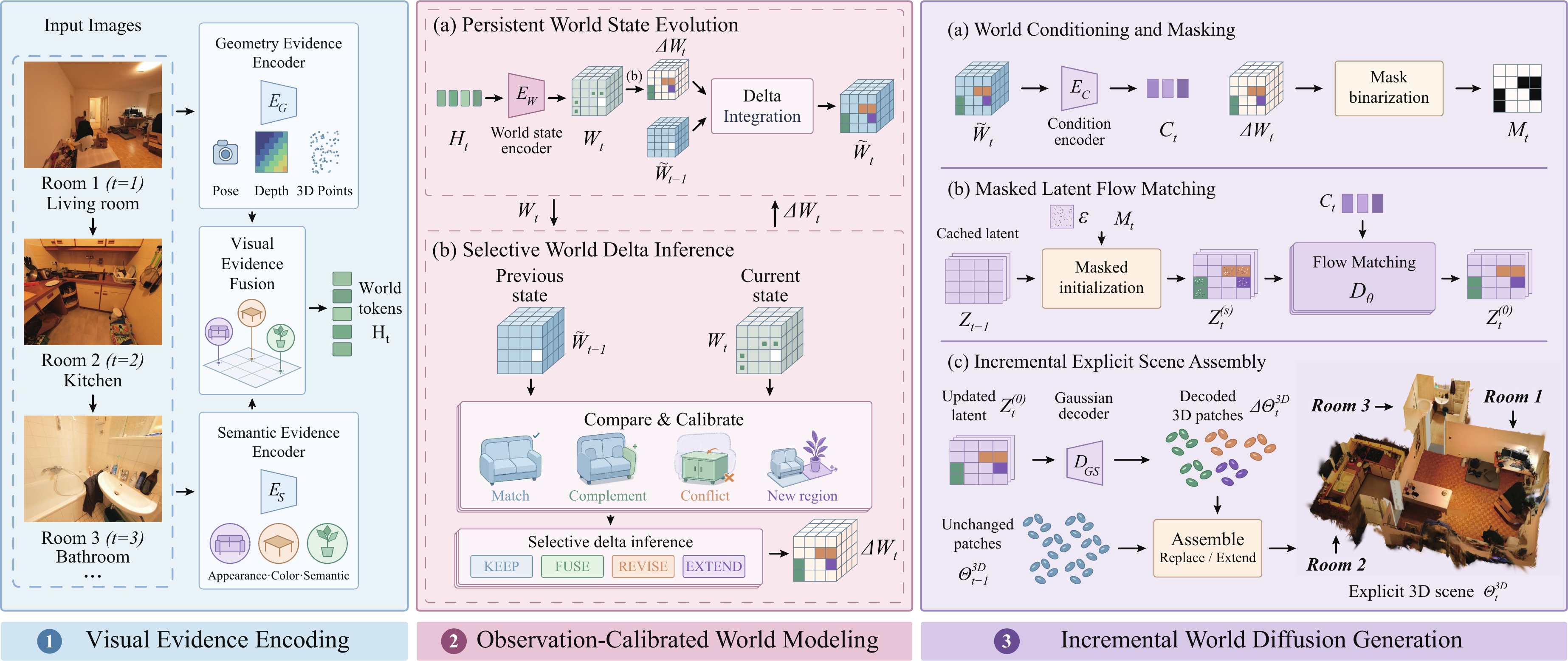}
	\caption{\textbf{Overview of World-$\delta$.}
		(1) \textbf{Visual Evidence Encoding} fuses geometric and semantic cues into spatially anchored world tokens.
		(2) \textbf{Observation-Calibrated World Modeling} compares evidence with a persistent state, using provenance, confidence,
		and uncertainty to infer selective deltas.
		(3) \textbf{Incremental World Diffusion Generation} uses the updated state to guide masked flow matching,
		freezes unaffected latents, and decodes only affected chunks for assembly with cached content.}
	\label{fig2}
\end{figure*}%
}

\SchedulePageFloats{2}{%
\begin{figure*}[!t]
	\centering
	\setlength{\abovecaptionskip}{0.1cm}
	\setlength{\belowcaptionskip}{-0.5cm}
	\includegraphics[width=\linewidth]{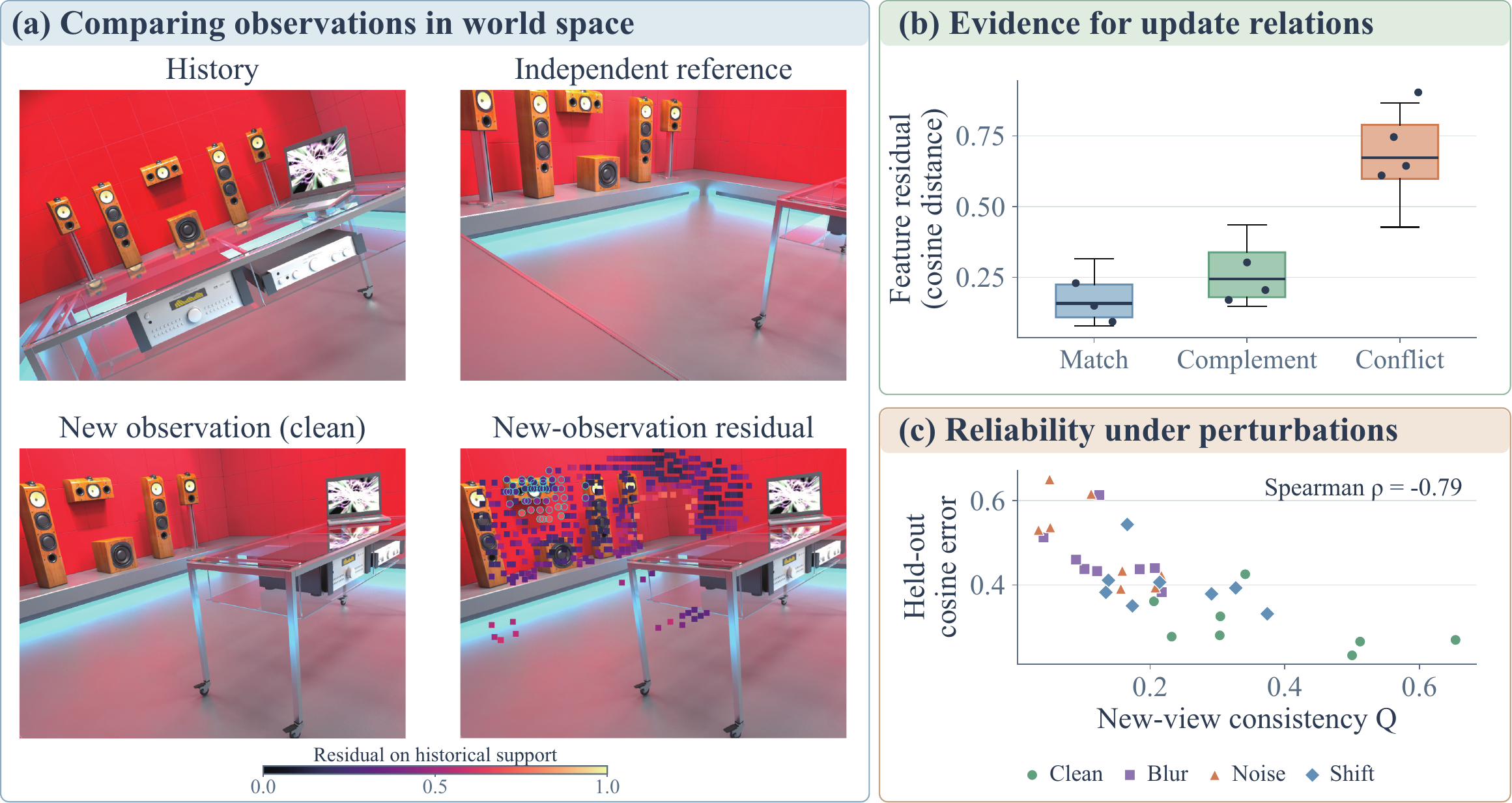}
	\caption{\textbf{Evidence reliability for observation-calibrated updates.}
		Using frozen DINOv2 features on four Hypersim scenes, we compare aligned
		observations and residuals on historical support (a), examine residual
		distributions for matching, complementary, and conflicting evidence (b),
		and relate new-view consistency to held-out feature error under
		perturbations (c). The analysis motivates considering content disagreement
		together with observation reliability before selecting world updates.}
	\label{fig3}
\end{figure*}%
}

\SchedulePageFloats{3}{%
\begin{figure*}[!t]
  \centering
  \setlength{\abovecaptionskip}{0.1cm}
  \setlength{\belowcaptionskip}{-0.5cm}
  \includegraphics[width=\linewidth]{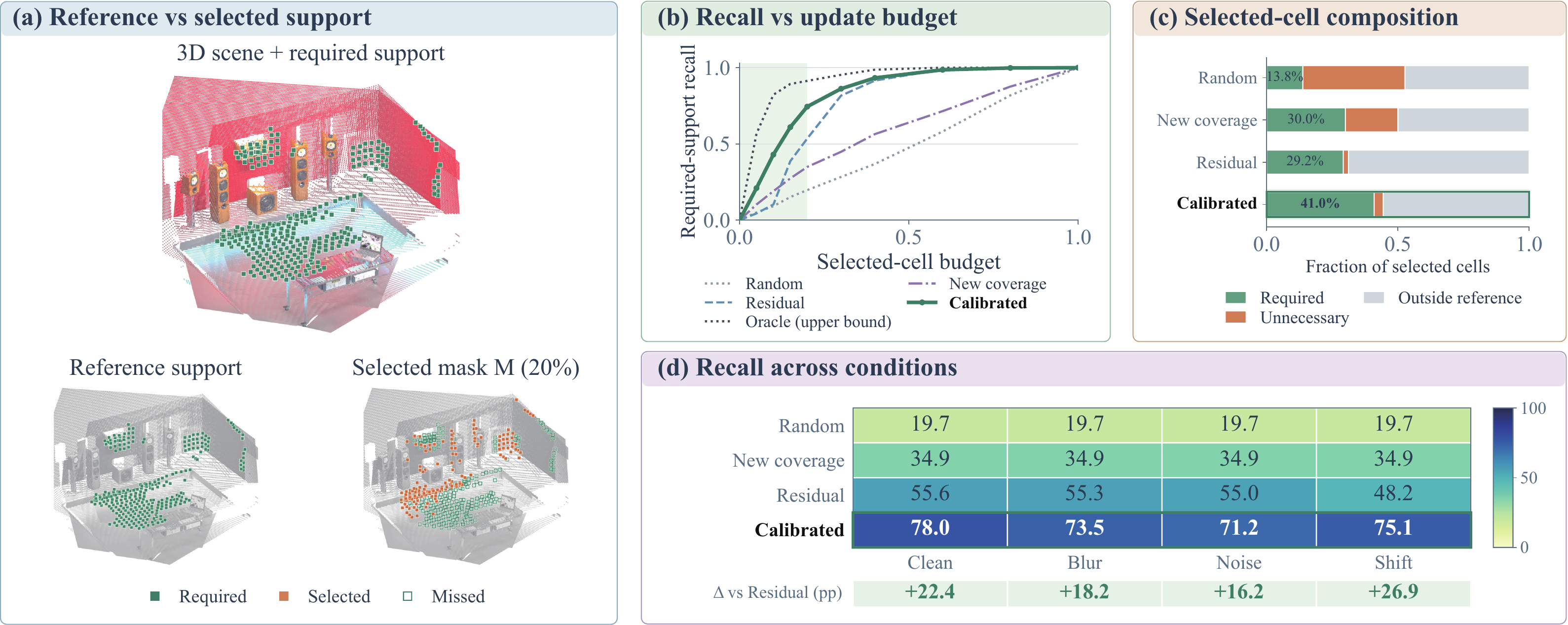}
  \caption{\textbf{Localization and allocation of incremental updates.}
  We compare calibrated, random, new-coverage, and residual-based selection under limited update budgets: reference support and
  a 20\% mask (a), recall versus budget (b), selected-cell composition (c),
  and recall under observation perturbations (d). Cells outside the reference support
  lack valid held-out evidence. Calibrated selection
  improves support recall at matched budgets, motivating delta-derived generation masks.}
  \label{fig4}
\end{figure*}%
}

\SchedulePageFloats{4}{%
\begin{table*}[!t]
\centering
\begin{minipage}[t]{\columnwidth}
\captionsetup{type=table}

\centering
\setlength{\abovecaptionskip}{0pt}
\setlength{\belowcaptionskip}{0.1cm}
\caption{Single-image 3D generation comparison.}
\label{tab:c1}
\begingroup
\setlength{\tabcolsep}{1.2pt}
\renewcommand{\arraystretch}{0.856}
\footnotesize
\addtolength{\aboverulesep}{-0.6026pt}
\addtolength{\belowrulesep}{-0.6026pt}
\begin{tabular*}{\linewidth}{@{\extracolsep{\fill}}lcccc@{}}
\toprule
\multirow{2}{*}{Method} & \multicolumn{4}{c}{RealEstate10K~\citep{re10k2018}} \\
\cmidrule(lr){2-5}
 & PSNR$\uparrow$ & SSIM$\uparrow$ & LPIPS$\downarrow$ & Time$\downarrow$ \\
\midrule
Lyra-1~\citep{lyra2026} & 21.8 & 0.749 & 0.221 & 75\,min \\
HY-World-2.0~\citep{hyworld2026} & 21.6 & 0.712 & 0.238 & 35\,min \\
3D-Belief~\citep{yin2026_3dbelief} & 16.4 & 0.593 & 0.339 & \textbf{35.6\,s} \\
InfiniSplat~\citep{infinisplat2026} & 18.8 & 0.698 & 0.291 & 68.3\,s \\
SplatAnything~\citep{yu2026splatanything} & 22.1 & 0.734 & 0.257 & 41.8\,s \\
\midrule
\textbf{Ours} & \textbf{27.5} & \textbf{0.812} & \textbf{0.183} & 59.2\,s \\
\midrule
\multirow{2}{*}{Method} & \multicolumn{4}{c}{Tanks and Temples~\citep{tanks2017}} \\
\cmidrule(lr){2-5}
 & PSNR$\uparrow$ & SSIM$\uparrow$ & LPIPS$\downarrow$ & Time$\downarrow$ \\
\midrule
Lyra-1~\citep{lyra2026} & 19.3 & 0.573 & 0.342 & 77\,min \\
HY-World-2.0~\citep{hyworld2026} & 20.5 & 0.694 & 0.307 & 36\,min \\
3D-Belief~\citep{yin2026_3dbelief} & 15.1 & 0.615 & 0.380 & \textbf{32.8\,s} \\
InfiniSplat~\citep{infinisplat2026} & 16.9 & 0.679 & 0.325 & 73.6\,s \\
SplatAnything~\citep{yu2026splatanything} & 19.6 & 0.708 & 0.311 & 44.7\,s \\
\midrule
\textbf{Ours} & \textbf{23.9} & \textbf{0.782} & \textbf{0.273} & 52.3\,s \\
\bottomrule
\end{tabular*}
\endgroup
\end{minipage}%
\hfill%
\begin{minipage}[t]{\columnwidth}
\captionsetup{type=table}

\centering
\setlength{\abovecaptionskip}{0pt}
\setlength{\belowcaptionskip}{0.1cm}
\caption{Continual RGB world update comparison.}
\label{tab:c2}
\begingroup
\setlength{\tabcolsep}{1.2pt}
\renewcommand{\arraystretch}{0.856}
\footnotesize
\addtolength{\aboverulesep}{-0.6026pt}
\addtolength{\belowrulesep}{-0.6026pt}
\begin{tabular*}{\linewidth}{@{\extracolsep{\fill}}llcccc@{}}
\toprule
\multirow{2}{*}{Type} & \multirow{2}{*}{Method} & \multicolumn{4}{c}{ScanNet-v2~\citep{scannet2017}} \\
\cmidrule(lr){3-6}
 & & PSNR$\uparrow$ & SSIM$\uparrow$ & LPIPS$\downarrow$ & Update$\downarrow$ \\
\midrule
\multirow{2}{*}{Recon.} & S3PO-GS~\citep{s3pogs2025} & 17.7 & 0.723 & 0.627 & 9.1\,s \\
 & OF3GS~\citep{of3gs2026} & 20.6 & 0.695 & 0.362 & \textbf{3.8\,s} \\
\midrule
\multirow{4}{*}{Gen.} & FlexWorld~\citep{chen2025flexworld} & 24.3 & 0.776 & 0.289 & 35\,min \\
 & Lyra-2~\citep{shen2026lyra2} & 23.5 & 0.795 & 0.263 & 32\,min \\
 & 3D-Belief~\citep{yin2026_3dbelief} & 18.4 & 0.672 & 0.548 & 19.7\,s \\
 & \textbf{Ours} & \textbf{28.7} & \textbf{0.841} & \textbf{0.192} & 6.1\,s \\
\midrule
\multirow{2}{*}{Type} & \multirow{2}{*}{Method} & \multicolumn{4}{c}{Replica~\citep{replica2019}} \\
\cmidrule(lr){3-6}
 & & PSNR$\uparrow$ & SSIM$\uparrow$ & LPIPS$\downarrow$ & Update$\downarrow$ \\
\midrule
\multirow{2}{*}{Recon.} & S3PO-GS~\citep{s3pogs2025} & 25.9 & 0.812 & 0.281 & 10.9\,s \\
 & OF3GS~\citep{of3gs2026} & 27.3 & 0.847 & 0.252 & \textbf{4.3\,s} \\
\midrule
\multirow{4}{*}{Gen.} & FlexWorld~\citep{chen2025flexworld} & 30.1 & 0.896 & 0.105 & 34\,min \\
 & Lyra-2~\citep{shen2026lyra2} & 28.6 & 0.914 & 0.133 & 33\,min \\
 & 3D-Belief~\citep{yin2026_3dbelief} & 25.5 & 0.829 & 0.291 & 22.4\,s \\
 & \textbf{Ours} & \textbf{34.8} & \textbf{0.948} & \textbf{0.079} & 5.2\,s \\
\bottomrule
\end{tabular*}
\endgroup
\end{minipage}%
\end{table*}
\begin{figure*}[!t]
\centering
\setlength{\abovecaptionskip}{0.1cm}
\setlength{\belowcaptionskip}{-0.5cm}
\includegraphics[width=\linewidth]{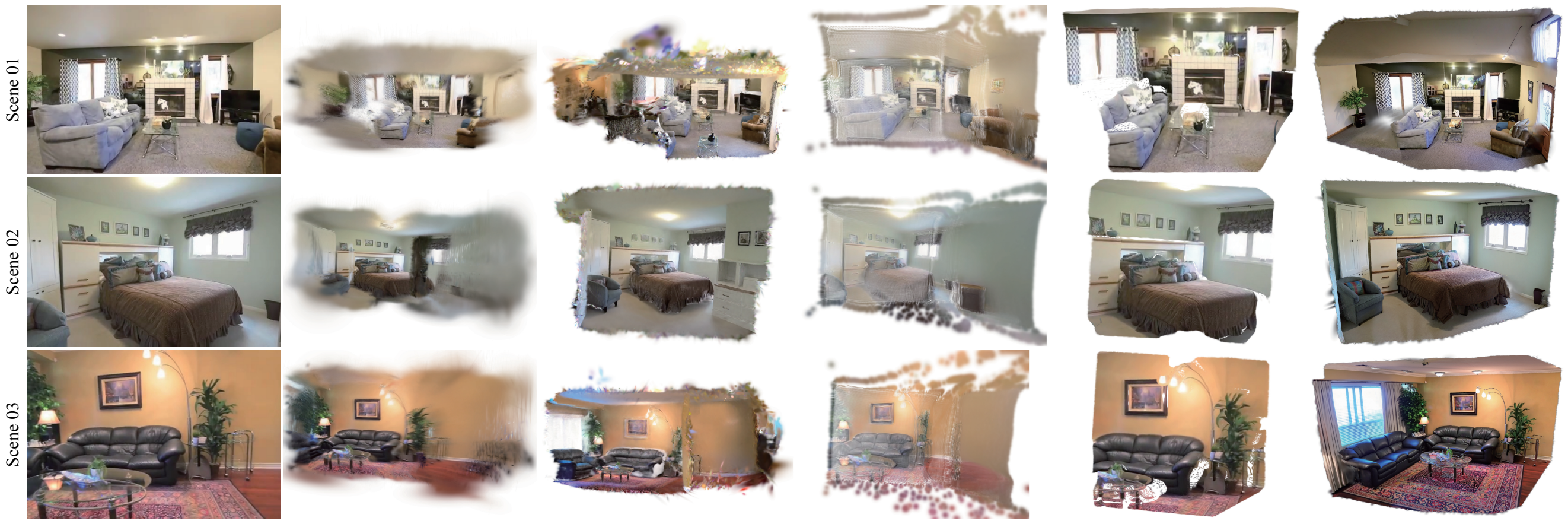}
\caption{Visualization of single-image 3D scene generation on RealEstate10K.}
\label{fig5}
\end{figure*}%
}

\SchedulePageFloats{5}{%
\begin{table*}[!t]
\centering
\begin{minipage}[t]{\columnwidth}
\captionsetup{type=table}
\centering
\setlength{\abovecaptionskip}{0cm}
\setlength{\belowcaptionskip}{0.1cm}
\caption{Visual evidence encoding ablation.}
\label{tab:a1}
\begingroup
\setlength{\tabcolsep}{1.2pt}
\renewcommand{\arraystretch}{2.0132}
\footnotesize
\begin{tabular*}{\linewidth}{@{\extracolsep{\fill}}lccccc@{}}
\toprule
Variant & LPIPS$\downarrow$ & \makecell{F-score\\@5cm$\uparrow$} & \makecell{Coverage\\@5cm$\uparrow$} & $G_{\rm corr}\uparrow$ & $F_{\rm keep}\downarrow$ \\
\midrule
w/o Semantic Features & 0.217 & 72.4\% & 79.8\% & 0.067 & 0.011 \\
w/o Geometry Features & 0.253 & 63.0\% & 71.6\% & 0.041 & 0.023 \\
\textbf{Full} & \textbf{0.192} & \textbf{74.3\%} & \textbf{80.8\%} & \textbf{0.081} & \textbf{0.008} \\
\bottomrule
\end{tabular*}
\endgroup
\end{minipage}%
\hfill%
\begin{minipage}[t]{\columnwidth}
\captionsetup{type=table}
\centering
\setlength{\abovecaptionskip}{0cm}
\setlength{\belowcaptionskip}{0.1cm}
\caption{Observation-calibrated world modeling ablation.}
\label{tab:a2}
\begingroup
\setlength{\tabcolsep}{1.2pt}
\renewcommand{\arraystretch}{1.45}
\fontsize{8}{9}\selectfont
\begin{tabular*}{\linewidth}{@{\extracolsep{\fill}}lcccc@{}}
\toprule
Variant & LPIPS$\downarrow$ & $E_{\rm corr}^{+}\downarrow$ & $G_{\rm corr}\uparrow$ & $F_{\rm keep}\downarrow$ \\
\midrule
w/o Provenance P & 0.223 & 0.116 & 0.044 & 0.018 \\
w/o Confidence Q & 0.215 & 0.104 & 0.056 & 0.025 \\
w/o Uncertainty U & 0.207 & 0.097 & 0.063 & 0.013 \\
Unconditional Fusion & 0.236 & 0.123 & 0.037 & 0.032 \\
\textbf{Full} & \textbf{0.192} & \textbf{0.079} & \textbf{0.081} & \textbf{0.008} \\
\bottomrule
\end{tabular*}
\endgroup
\end{minipage}%
\end{table*}
\begin{figure*}[!t]
\centering
\setlength{\abovecaptionskip}{0.1cm}
\setlength{\belowcaptionskip}{-0.5cm}
\includegraphics[width=\linewidth]{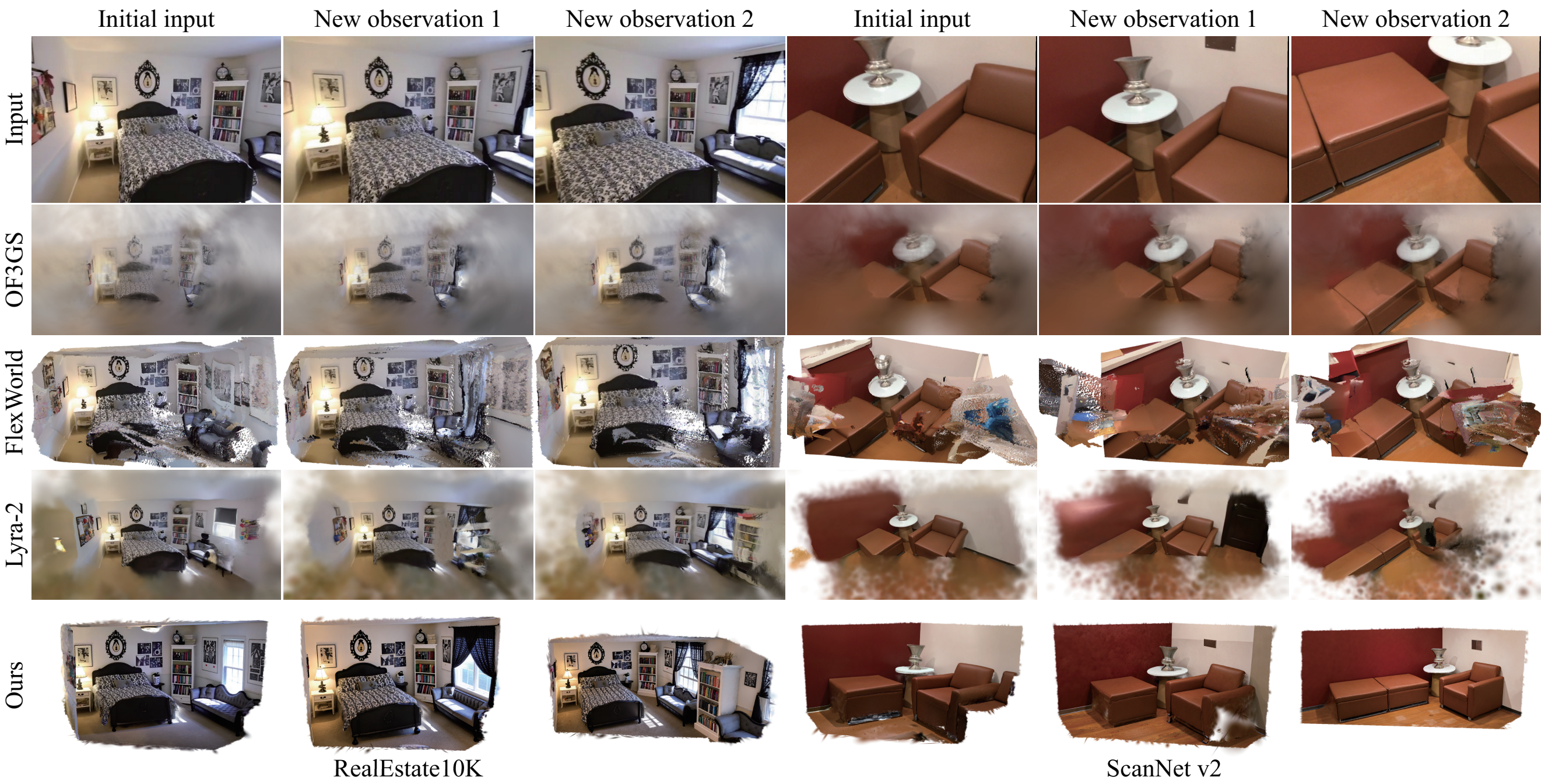}
\caption{Visualization of continual 3D world updates on RealEstate10K and ScanNet-v2.}
\label{fig6}
\end{figure*}%
}

\SchedulePageFloats{6}{%
\begin{table*}[!t]
\centering
\begin{minipage}[t]{\columnwidth}
\captionsetup{type=table}
\centering
\setlength{\abovecaptionskip}{0cm}
\setlength{\belowcaptionskip}{0.1cm}
\caption{Incremental world diffusion generation ablation.}
\label{tab:a3}
\begingroup
\setlength{\tabcolsep}{1.2pt}
\renewcommand{\arraystretch}{1.8058}
\fontsize{7}{8}\selectfont
\begin{tabular*}{\linewidth}{@{\extracolsep{\fill}}lcccccc@{}}
\toprule
Variant & LPIPS$\downarrow$ & $G_{\rm corr}\uparrow$ & $F_{\rm keep}\downarrow$ & \makecell{Update\\(s)$\downarrow$} & \makecell{Decode\\(s)$\downarrow$} & \makecell{GPU\\(GB)$\downarrow$} \\
\midrule
\makecell[l]{Visibility-only\\Mask} & 0.247 & 0.026 & \textbf{0.007} & \textbf{5.0} & \textbf{1.1} & \textbf{11.1} \\
Global Diffusion & \textbf{0.189} & \textbf{0.085} & 0.042 & 24.3 & 12.7 & 16.8 \\
\makecell[l]{w/o Hard\\Freezing} & 0.214 & 0.082 & 0.034 & 25.6 & 12.7 & 16.8 \\
Full Decoding & 0.192 & 0.081 & 0.008 & 16.7 & 12.6 & 16.4 \\
\textbf{Full} & 0.192 & 0.081 & 0.008 & 5.2 & 1.3 & 11.2 \\
\bottomrule
\end{tabular*}
\endgroup
\end{minipage}%
\hfill%
\begin{minipage}[t]{\columnwidth}
\captionsetup{type=table}
\centering
\setlength{\abovecaptionskip}{0cm}
\setlength{\belowcaptionskip}{0.1cm}
\caption{Correction and retention under new evidence.}
\label{tab:d1}
\begingroup
\setlength{\tabcolsep}{1.2pt}
\renewcommand{\arraystretch}{0.78265}
\footnotesize
\addtolength{\aboverulesep}{-0.9877pt}
\addtolength{\belowrulesep}{-0.9877pt}
\begin{tabular*}{\linewidth}{@{\extracolsep{\fill}}lcccc@{}}
\toprule
\multirow{2}{*}{Variant} & \multicolumn{4}{c}{ScanNet v2~\citep{scannet2017}} \\
\cmidrule(lr){2-5}
 & $E_{\rm corr}^{+}\downarrow$ & $G_{\rm corr}\uparrow$ & $E_{\rm keep}^{+}\downarrow$ & $F_{\rm keep}\downarrow$ \\
\midrule
Extend-only & 0.148 & 0.012 & \textbf{0.083} & \textbf{0.003} \\
Batch-Rebuild & \textbf{0.075} & \textbf{0.085} & 0.115 & 0.035 \\
\textbf{Full} & 0.079 & 0.081 & 0.088 & 0.008 \\
\midrule
\multirow{2}{*}{Variant} & \multicolumn{4}{c}{Replica~\citep{replica2019}} \\
\cmidrule(lr){2-5}
 & $E_{\rm corr}^{+}\downarrow$ & $G_{\rm corr}\uparrow$ & $E_{\rm keep}^{+}\downarrow$ & $F_{\rm keep}\downarrow$ \\
\midrule
Extend-only & 0.112 & 0.008 & \textbf{0.053} & \textbf{0.001} \\
Batch-Rebuild & \textbf{0.049} & \textbf{0.071} & 0.083 & 0.031 \\
\textbf{Full} & 0.054 & 0.066 & 0.058 & 0.006 \\
\bottomrule
\end{tabular*}
\endgroup
\end{minipage}%
\end{table*}
\begin{figure*}[!t]
  \centering
  \setlength{\abovecaptionskip}{0.1cm}
  \setlength{\belowcaptionskip}{-0.5cm}
  \includegraphics[width=\linewidth]{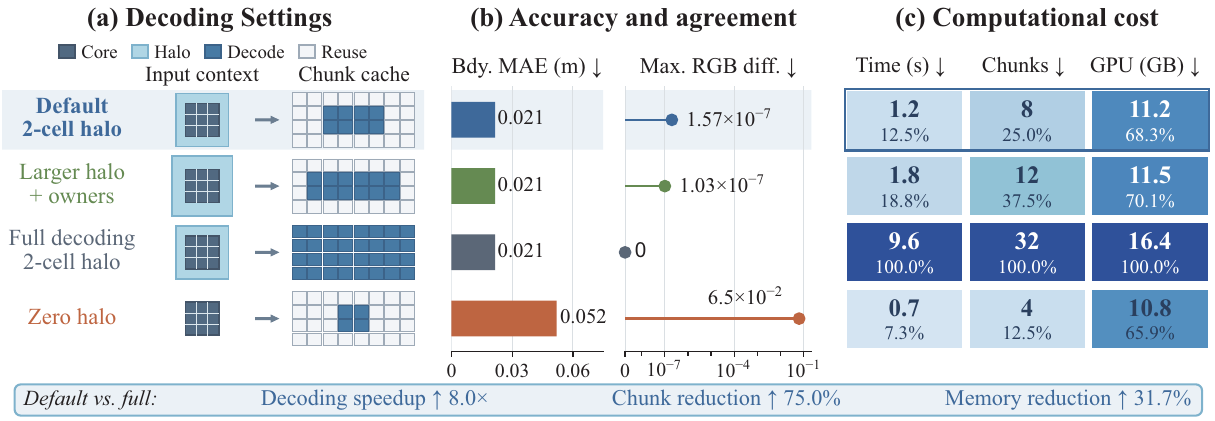}
  \caption{\textbf{Decoding context, boundary agreement, and computational cost.}
  (a) Decoder neighborhoods distinguish core cells, read-only context, and chunks eligible for cache reuse.
  (b) Boundary depth error and RGB deviation reveal the effect of truncating context.
  (c) Time, chunk count, and GPU memory expose the cost of larger neighborhoods relative to full decoding.}
  \label{fig7}
\end{figure*}%
}

\SchedulePageFloats{7}{%
\begin{figure*}[!t]
  \centering
  \setlength{\abovecaptionskip}{0.1cm}
  \setlength{\belowcaptionskip}{-0.5cm}
  \includegraphics[width=\linewidth]{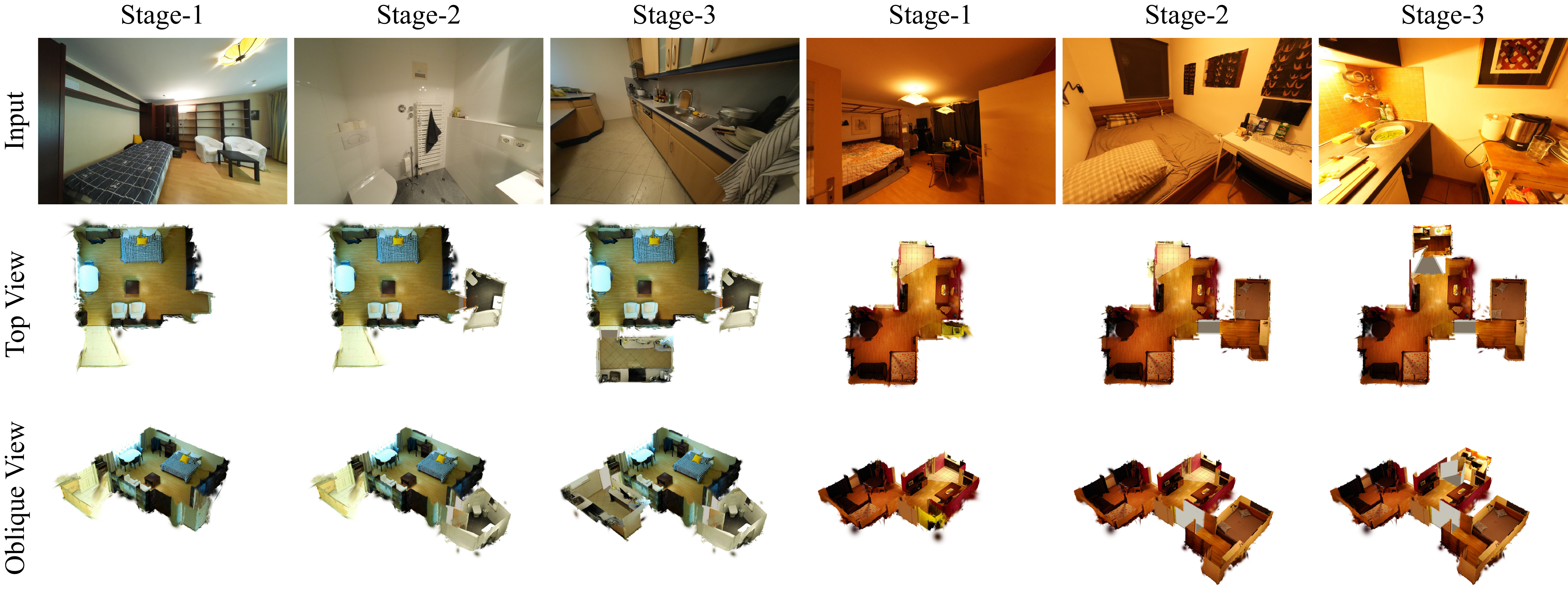}
  \caption{\textbf{Cross-room incremental scene expansion.}
  Successive observations reveal different rooms in two indoor scenes of \mbox{ScanNet++}~\citep{yeshwanth2023scannetpp}.
  Top and oblique views show how World-$\delta$ integrates newly observed rooms
  into the evolving scene.}
  \label{fig8}
\end{figure*}%
}

\SchedulePageFloats{8}{%
\begin{table}[!t]
\centering
\setlength{\abovecaptionskip}{0cm}
\setlength{\belowcaptionskip}{0.1cm}
\caption{Selection under a matched update budget.}
\label{tab:d2_budget}
\begingroup
\setlength{\tabcolsep}{1.2pt}
\renewcommand{\arraystretch}{1.12}
\fontsize{8}{9}\selectfont
\begin{tabular*}{\linewidth}{@{\extracolsep{\fill}}lcccc@{}}
\toprule
Selection rule & LPIPS$\downarrow$ & $G_{\rm corr}\uparrow$ & $F_{\rm keep}\downarrow$ & Update(s)$\downarrow$ \\
\midrule
w/o U in Selection & 0.216 & 0.061 & 0.014 & 5.7 \\
Residual-only & 0.227 & 0.059 & 0.019 & 6.8 \\
Visibility-only & 0.249 & 0.026 & \textbf{0.007} & \textbf{5.0} \\
Random & 0.263 & 0.019 & 0.026 & 9.3 \\
\textbf{Calibrated Delta} & \textbf{0.192} & \textbf{0.081} & 0.008 & 5.2 \\
\bottomrule
\end{tabular*}
\endgroup
\end{table}%
}

\begin{abstract}
Existing 3D generation methods synthesize visually compelling scenes from fixed inputs, yet struggle to continually expand and revise a 3D world as new observations arrive. Iterative approaches often fuse real observations and generated content into a single representation, treating generated content as established structure rather than revisable predictions. Consequently, early generation errors can persist and accumulate despite new evidence. We propose \mbox{World-\(\delta\)}, an incremental 3D scene generation framework combining uncertainty-aware persistent world modeling with delta-guided diffusion for continual scene expansion and correction. Our framework comprises three components:
1) Visual Evidence Encoding integrates geometric and semantic cues from incoming images into unified world tokens.
2) Observation-Calibrated World Modeling maintains a persistent latent world state that distinguishes observation-supported content from provisional completion hypotheses. New evidence drives selective world deltas to revise conflicting hypotheses while preserving well-supported content.
3) Incremental World Diffusion Generation uses delta-derived masks to restrict diffusion to regions requiring updates, freezing unchanged regions. Only affected 3D chunks are decoded and integrated into the existing scene.
Together, these components enable continual world expansion and revision while reducing redundant generation and preserving scene consistency. Experiments demonstrate the effectiveness of \mbox{World-\(\delta\)} in incremental 3D scene generation, observation-driven correction, scene consistency, and update efficiency over successive scene updates.
\end{abstract}
\vspace{-12pt} 

\section{Introduction}
\label{sec:introduction}

3D scene generation synthesizes worlds with plausible geometry and appearance from input conditions. Despite compelling results, existing methods struggle to expand and revise worlds as observations accumulate. This limitation stems from inadequate modeling of how new observations relate to the existing generated scene.

Typically, iterative methods relate new observations to the current scene through geometric alignment and fusion, integrating them into a shared 3D representation~\citep{fridman2023scenescape,chung2025luciddreamer}. However, without distinguishing information sources and reliability, early errors may be reinforced, leaving unclear what to preserve or revise as new evidence arrives. Thus, the key to incremental 3D generation is to reconcile new evidence with the existing world, enabling continual expansion and correction while preserving reliable content.

Consequently, we identify two insights. First, generated hypotheses should remain revisable as evidence evolves. Tracking provenance and confidence in a persistent world state allows new observations to confirm, revise, or reject hypotheses while preserving well-supported content. Second, generation should follow selective world-state changes. Since observations may reveal, correct, or merely confirm scene content, inferring state deltas before generation focuses computation on regions requiring updates.

In this work, we introduce \mbox{World-\(\delta\)}, an incremental 3D scene generation framework for continuous observations. Visual Evidence Encoding fuses geometric and semantic cues into world tokens. Observation-Calibrated World Modeling maintains an uncertainty-aware persistent state, inferring selective deltas to integrate new evidence and revise conflicting hypotheses while preserving supported content. Incremental World Diffusion Generation uses these deltas to mask diffusion conditioned on the updated state, freezing unaffected latents. Only changed 3D chunks are decoded and integrated into the existing scene, enabling efficient expansion and correction. Our main contributions to incremental 3D generation are summarized as follows:
\begin{itemize}[leftmargin=*]
  \item We introduce \mbox{World-\(\delta\)}, a framework that continually expands and corrects a 3D world from incoming RGB observations, integrating evidence while preserving reliable history and keeping provisional completions revisable.
  \item Observation-Calibrated World Modeling maintains an uncertainty-aware persistent state, comparing new evidence with history to infer selective updates that revise conflicting hypotheses while retaining supported content.
  \item Incremental World Diffusion Generation uses state deltas to guide selective diffusion and decode only affected chunks, preserving unchanged content and reducing redundant computation during continual world updates.
  \item We aggregate and adapt large-scale Gaussian datasets for incremental 3D Gaussian generation. Experiments show improved rendering quality while balancing observation-driven correction, historical retention, and update cost.
\end{itemize}

\section{Related Work}
\label{sec:related_works}

\textbf{3D Generation.}
Recent advances in 3D and 4D generation combine structured scene representations with diffusion priors~\citep{tang2024dreamgaussian,chen2024gsgen,yi2024gaussiandreamer,li2025DreamScene,lin2025diffsplat,lyra2026,hyworld2026,zhou2025uni4d,wang2026cog2gen3d,wang2026st}. Lyra-1~\citep{lyra2026} distills video diffusion priors into explicit Gaussian scenes, whereas HY-World-2.0~\citep{hyworld2026} combines panorama generation, view expansion, and Gaussian scene composition to construct complete worlds. Despite producing high-quality results, these methods offer limited support for observations arriving after initial generation. Our work aims to integrate such evidence into existing scenes to enable coherent world expansion.

\noindent
\textbf{Persistent World Modeling.}
Existing methods encode spatial and spatiotemporal context in learned scene representations~\citep{wang2025cut3r,garcin2026persist,wang20264d,zhou2026llava} or retain observation history through retrieval and explicit spatial memories~\citep{xiao2025worldmem,shen2026lyra2,zhao2026spatia,lv2026octworld}. 3D-Belief~\citep{yin2026_3dbelief} maintains explicit beliefs about unobserved regions and updates them with new observations. These approaches improve spatial consistency, but retaining history alone does not specify how to reconcile new evidence with prior hypotheses or localize updates. We calibrate persistent states using provenance, observation confidence, and generative uncertainty to distinguish supported content from provisional completions and infer selective updates.

\noindent
\textbf{Incremental Diffusion Generation.}
Incremental 3D diffusion must update scenes while preserving unaffected content. Existing pipelines generate complete samples under fixed conditioning~\citep{lin2025diffsplat} or complete partial scenes through view-space inpainting and fusion or pixel-space diffusion~\citep{lei2023rgbd2,chung2025luciddreamer,chen2025flexworld,yu2025wonderworld,hoellein2023text2room,gao2026pixworld}. Hole filling can miss errors in occupied regions, while broad resampling risks disturbing reliable history and repeating computation. World-$\delta$ aligns diffusion updates with observation-calibrated deltas, freezes unaffected latents, and decodes only affected chunks for efficient, targeted correction.

\section{Method}
\label{sec:method}

\subsection{Overview}
World-$\delta$ expands and corrects a 3D world from RGB observations $\mathcal I_t=\{I_{t,j}\}_{j=1}^{N_t}$, where $t$ indexes observation rounds. Fig.~\ref{fig2} summarizes three stages.

\noindent
\textbf{Visual Evidence Encoding} fuses geometric and semantic features, $H_t^G$ and $H_t^S$, into world tokens using predicted poses $\Pi_t$, depths $D_t$, 3D points $X_t^{3D}$, and geometric reliability $Q_t^G$ to obtain spatially grounded tokens:
\begin{equation}
\label{eq:visual_fusion}
H_t=F_{\mathrm{vis}}(H_t^G,H_t^S;\Pi_t,D_t,X_t^{3D},Q_t^G).
\end{equation}
\textbf{Observation-Calibrated World Modeling} encodes current evidence as $\mathcal W_t$, compares it with history $\widetilde{\mathcal W}_{t-1}$, and integrates a selective delta into the persistent state:
\begin{equation}
\label{eq:state_update}
\begin{gathered}
\Delta\mathcal W_t=F_\Delta(\widetilde{\mathcal W}_{t-1},\mathcal W_t), \\
\widetilde{\mathcal W}_t=F_{\mathrm{state}}(\widetilde{\mathcal W}_{t-1},\mathcal W_t,\Delta\mathcal W_t),\quad t\geq1.
\end{gathered}
\end{equation}
\noindent
\textbf{Incremental World Diffusion Generation} conditions masked flow matching on the integrated state, refreshes selected latents as $Z_t=Z_t^{(0)}$, and decodes affected chunks into $\Theta_t^{3D}$, with unaffected chunks retained from history.

\subsection{Visual Evidence Encoding}
\textbf{Geometry and Semantic Evidence Encoders.}
LingBot-Map~\citep{chen2026lingbotmap} yields $(H_t^G,\Pi_t,D_t,X_t^{3D},Q_t^G)=E_G(\mathcal I_t)$, with back-projected anchors and depth confidence calibrated by visibility and geometric consistency. Frozen DINOv2 ViT-B/14~\citep{dosovitskiy2021vit,oquab2023dinov2} supplies appearance and semantic features $H_t^S=E_S(\mathcal I_t)$ alongside the available geometric features.

\noindent
\textbf{Visual Evidence Fusion.}
$F_{\mathrm{vis}}$ fuses geometry and semantics at 3D anchors into $H_t=\{(h_{t,n},a_{t,n},p_{t,n},q_{t,n}^{\mathrm{obs}})\}_{n=1}^{L_t}$: features, anchors, image/round provenance, and confidence.

\subsection{Observation-Calibrated World Modeling}
A geometry-assisted probe on four Hypersim scenes~\citep{roberts2021hypersim} with frozen DINOv2 yields larger residuals for controlled history conflicts (Fig.~\ref{fig3}(b)). Higher new-view consistency $Q$ corresponds to lower held-out error (Fig.~\ref{fig3}(c)). Both cues jointly guide evidence-driven update selection.

\noindent
\textbf{Persistent World State Evolution.}
For $t\geq1$, $E_W$ forms $\mathcal W_t=E_W(H_t;G_{t-1})$ in persistent coordinates. After generation, the integrated state stores generative latents $Z_t$, topology and accepted evidence attributes $G_t$, provenance $P_t$, observation confidence $Q_t$, and generative uncertainty $U_t$ within the same persistent world representation:
\begin{equation}
\label{eq:world_states}
\widetilde{\mathcal W}_t=(Z_t,G_t,P_t,Q_t,U_t).
\end{equation}
Confidence measures observational support. Uncertainty estimates generation error. At $t=0$, $\mathcal W_0=E_W(H_0)$ establishes the coordinate frame, and $(\widetilde{\mathcal W}_0,\Omega_0^\Delta)=F_{\mathrm{init}}(\mathcal W_0)$ marks all allocated locations for generation. Later rounds integrate each observation once via Eq.~(\ref{eq:state_update}). Generation refreshes latents in the same state, which becomes subsequent history as new observations arrive in later update rounds.

\noindent
\textbf{Selective World Delta Inference.} We compare evidence with historical world state using visibility, provenance, confidence, and uncertainty to assign $A_t(v)\in\{\mathrm{KEEP},\mathrm{FUSE},\mathrm{REVISE},\mathrm{EXTEND}\}$. KEEP preserves content, FUSE integrates compatible evidence, REVISE corrects conflicting content, and EXTEND allocates new regions. Reliable observations can revise uncertain completions. Weak or occluded evidence is downweighted. The structured edit is formulated as:
\begin{equation}
\label{eq:world_delta}
\Delta\mathcal W_t=(\Delta Z_t,\Delta G_t,\Delta P_t,\Delta Q_t,\Delta U_t,\Omega_t^\Delta).
\end{equation}
The delta inserts, replaces, or removes content. $\Delta Z_t$ requests regeneration. $\Omega_t^\Delta$ marks content and topology edits requiring realization. Metadata-only fusion skips diffusion.

\subsection{Incremental World Diffusion Generation}
Fig.~\ref{fig4}(b)--(d) shows that calibrated selection better recovers required update support than coverage-only or residual-only selection at limited budgets. We therefore derive editable regions from calibrated deltas and update chunks according to decoding dependencies while retaining required context.

\noindent
\textbf{World-State Conditioning and Update Masking.}
Conditioning uses $C_t=E_C(\widetilde{\mathcal W}_t)$. We form $M_t=R_{\mathrm{mask}}(\Omega_t^\Delta;\mathcal L_{\mathrm{diff}})$ on the diffusion lattice, including required boundary cells. Across channels, $M_t(v)=1$ permits writes. Zero preserves history. New slots and latent changes are masked. Neighboring latents provide read-only context. Metadata-only edits can bypass diffusion.

\noindent
\textbf{Masked Latent Initialization.}
We align history to current topology as $Z_t^{\mathrm{ref}}=\operatorname{Align}(Z_{t-1};G_t)$, preserving retained values. Let $s$ be the number of integration steps and $k=s,\ldots,0$ index sampling independently of $t$. Masked cells start from Gaussian noise for local regeneration:
\begin{equation}
\label{eq:diffusion_initialization}
Z_t^{(s)}=(1-M_t)\odot Z_t^{\mathrm{ref}}+M_t\odot\epsilon_t,\qquad\epsilon_t\sim\mathcal N(0,I).
\end{equation}
New reference entries are zero-initialized and masked, so noise replaces them. At initialization, $Z_0^{\mathrm{ref}}=0$, $M_0=1$ on the allocated domain, and $C_0$ supplies observation conditioning from the initially available input RGB images.

\noindent
\textbf{Selective Latent Update with Flow Matching.}
We train $D_\theta$ by flow matching~\citep{lipman2023flowmatching}. For a clean target $Z^\star$ and $\tau\in[0,1]$, editable cells follow $(1-\tau)Z^\star+\tau\epsilon$ with target velocity $\epsilon-Z^\star$. Others retain history. Sampling uses $\tau_k=k/s$ for $k=s,\ldots,1$, from noise to clean latents:
\begin{equation}
\label{eq:masked_diffusion}
\begin{gathered}
V_t^{(k)}=D_\theta(Z_t^{(k)},\tau_k\mid C_t,M_t,Z_t^{\mathrm{ref}}), \\
Z_t^{(k-1)}=(1-M_t)\odot Z_t^{\mathrm{ref}}+
M_t\odot\left[Z_t^{(k)}-\frac{1}{s}V_t^{(k)}\right].
\end{gathered}
\end{equation}
State latents become $Z_t=Z_t^{(0)}$. Provenance and uncertainty record realizations. Topology and confidence stay calibrated. Generation adds no observational support.

\noindent
\textbf{Incremental Explicit Scene Assembly.}
Our Gaussian decoder maps spatially anchored latent chunks to explicit Gaussian patches under the current scene topology. Latent changes and topology edits determine which chunks require decoding, with neighboring cells providing read-only context. Only selected chunks write their outputs, while unaffected patches are reused from the cache. This separates decoding context from writable content, enabling local correction and expansion while preserving reliable history.

Let $\mathscr C_t$ index chunks in $G_t$. Their decoding dependencies $\mathcal R_c$ cover neighborhoods and topology before and after edits. Affected chunks satisfy
\begin{equation}
\label{eq:affected_chunks}
\begin{split}
\mathscr C_t^\Delta=\bigl\{c\in\mathscr C_{t-1}\cup\mathscr C_t:\quad &\\
\mathcal R_c\cap\bigl(\Omega_t^\Delta\cup\operatorname{WorldSupp}(M_t)\bigr)&\neq\varnothing\bigr\},\quad t\geq1.
\end{split}
\end{equation}
$\operatorname{WorldSupp}$ maps the mask to world coordinates. $\mathscr C_0^\Delta=\mathscr C_0$ initializes all chunks. With $Z_{t,c}$ and required context, $D_{GS}$ decodes affected chunks under the current topology:
\begin{equation}
\label{eq:explicit_assembly}
\begin{gathered}
\Delta\Theta_t^{3D}=\{D_{GS}(Z_{t,c},G_t):c\in\mathscr C_t^\Delta\cap\mathscr C_t\}, \\
\Theta_t^{3D}=\operatorname{Assemble}(\Theta_{t-1}^{3D},\Delta\Theta_t^{3D};\mathscr C_t^\Delta).
\end{gathered}
\end{equation}
Starting from $\Theta_{-1}^{3D}=\varnothing$, assembly replaces or adds affected patches, removes $\mathscr C_t^\Delta\setminus\mathscr C_t$, and reuses unaffected entries. Decoding reads a two-cell halo and writes only selected owners' outputs, with other chunks unchanged.

\subsection{Training Objectives}
\label{sec:objectives}
\noindent
\textbf{Gaussian Reconstruction Loss.}
The codec uses reconstruction and a standard normal prior:
\begin{equation}
\label{eq:codec_loss}
\mathcal L_{\mathrm{codec}}=\mathcal L_{\mathrm{slot}}+\mathcal L_{\mathrm{render}}+\beta_{\mathrm{KL}}\mathcal L_{\mathrm{KL}}.
\end{equation}
$\mathcal L_{\mathrm{slot}}$ supervises occupancy and attributes. $\mathcal L_{\mathrm{render}}$ combines RGB $L_1$, SSIM, depth, and coverage.

\noindent
\textbf{State Calibration Loss.}
State learning combines weighted action cross-entropy, binary cross-entropy for reliability and free space, squared uncertainty error, and feature cosine distance for calibrating the persistent world-state attributes:
\begin{equation}
\label{eq:state_loss}
\mathcal L_{\mathrm{state}}=\mathcal L_{\mathrm{action}}+\mathcal L_q+\mathcal L_U+0.2\mathcal L_{\mathrm{feature}}+0.5\mathcal L_{\mathrm{free}}.
\end{equation}

\noindent
\textbf{Masked Flow-Matching Loss.}
On the linear path above, $V_\theta$ is the predicted velocity and $T_{\mathrm{valid}}$ marks valid targets. Using $S_t=M_t\odot T_{\mathrm{valid}}$, we minimize
\begin{equation}
\label{eq:flow_loss}
\mathcal L_{\mathrm{FM}}=\mathbb E_{\tau,\epsilon}\!\left[\frac{\|S_t\odot(V_\theta-(\epsilon-Z^\star))\|_2^2}{d_z\sum_v S_t(v)+10^{-8}}\right],\qquad d_z=64.
\end{equation}
We average over valid targets. With the codec frozen, we use $0.05\mathcal L_{\mathrm{render}}$ to decoded clean-latent estimates every ten steps. Target validity selects supervision targets.

\section{Experiments}
\label{sec:experiments}

\subsection{Implementation Details}
\label{sec:implementation}
\noindent
\textbf{Network configuration.}
Evidence is encoded as 256-dimensional world tokens. Each chunk has $8^3$ 64-channel latent cells and up to 4,096 Gaussians. A six-block Transformer predicts velocities for 50 Euler steps. The Gaussian decoder uses two valid $3\times3\times3$ convolutions, giving a two-cell dependency radius on the latent grid. Each selected output chunk is decoded with a two-cell halo that supplies the required neighboring features. Only the chunk's owned Gaussian outputs are retained. Halo cells provide context without extending the writable region. This bounded dependency determines the affected-chunk selection in Eq.~(\ref{eq:affected_chunks}) and supports reuse of unaffected decoded patches.

\noindent
\textbf{Training data.}
We aggregate training splits from SceneSplat-49K~\citep{ma2025scenesplatpp}: 3RScanGS, ARKitScenesGS, Matterport3DGS, HyperSimGS, ReplicaGS, ScanNetPPGS-V1/V2, ScanNetGS, DL3DV-GS, HoliCityGS, AriaSyntheticEnvGS, and crowdsourced scenes. We normalize coordinates and convert scenes into canonical chunk targets, pairing RGB prefixes with Gaussian and latent supervision for initialization, expansion, and revisiting. Reference Gaussians and available geometric labels provide offline supervision. Inference uses only RGB prefixes.

\noindent
\textbf{Training configuration.}
Training covers codec, state, flow, and sequential calibration stages. Non-codec stages use AdamW with learning rate $10^{-4}$, weight decay 0.01, and 16-chunk batches. Adaptation uses $10^{-5}$. Evidence backbones remain frozen. Codec pretraining ramps $\beta_{\mathrm{KL}}$ to $10^{-4}$ during the codec's Gaussian reconstruction training stage.

\subsection{Comparison Experiments}
\label{sec:comparisons}
We evaluate single-image generation and continual RGB updates with identical observations and query views. All compared methods use the same training and testing configurations, including training data, computational resources, and evaluation environments. All training is conducted on eight NVIDIA A100 GPUs, and all inference runs on a single NVIDIA A100 GPU under the same test setup.

\noindent
\textbf{3D Generation Comparison.}
We compare with Lyra-1~\citep{lyra2026}, HY-World-2.0~\citep{hyworld2026}, 3D-Belief~\citep{yin2026_3dbelief}, InfiniSplat~\citep{infinisplat2026}, and SplatAnything~\citep{yu2026splatanything} on RealEstate10K~\citep{re10k2018} and Tanks and Temples~\citep{tanks2017} to evaluate rendering quality and generation cost from a single RGB image. As shown in Table~\ref{tab:c1}, World-$\delta$ achieves favorable overall performance across the two datasets, with clear advantages on most of the reported measures. Fig.~\ref{fig5} presents qualitative comparisons on DL3DV~\citep{dl3dv2024}, RealEstate10K, and ScanNet v2~\citep{scannet2017}. While the baselines often leave substantial holes in unseen regions, World-$\delta$ uses learned priors to generate more complete scenes, illustrating scene generation and completion across varied indoor layouts from the available RGB inputs.

\noindent
\textbf{3D Incremental Generation Comparison.}
We compare reconstruction baselines S3PO-GS~\citep{s3pogs2025} and OF$^3$GS~\citep{of3gs2026}, and generation baselines FlexWorld~\citep{chen2025flexworld}, Lyra2~\citep{shen2026lyra2}, and 3D-Belief~\citep{yin2026_3dbelief} on ScanNet v2~\citep{scannet2017} and Replica~\citep{replica2019}. All methods receive the same RGB observation prefixes and are evaluated at shared query views after successive updates. As shown in Table~\ref{tab:c2}, World-$\delta$ shows consistent advantages over the reconstruction and generation baselines across the reported comparisons on both datasets. Fig.~\ref{fig6} illustrates scene expansion and revision as new observations arrive. Together, these results support the use of selective world updates to integrate new evidence with favorable rendering quality and update cost over successive update rounds.

\subsection{Ablation Study}
\label{sec:ablations}
We ablate three components on ScanNet v2 using LPIPS, surface F-score and coverage at 5\,cm, correction gain $G_{\rm corr}=E_{\rm corr}^{-}-E_{\rm corr}^{+}$, and retention error increase $F_{\rm keep}=E_{\rm keep}^{+}-E_{\rm keep}^{-}$. Superscripts $-$/$+$ mark before/after updates. Correction uses common newly revealed support. Retention uses reliable historical support. We average relative depth errors clipped at one. Missing predictions count as one. Both metrics are absolute differences. Higher $G_{\rm corr}$ and lower $F_{\rm keep}$ are better under this evaluation protocol.

\noindent
\textbf{Effectiveness of Visual Evidence Encoding.}
In Table~\ref{tab:a1}, we ablate geometric and semantic features to assess their roles in world updates. Removing semantics degrades perceptual quality and correction. Removing geometry causes larger losses in reconstruction accuracy and coverage.

\noindent
\textbf{Superiority of Observation-Calibrated World Modeling.}
In Table~\ref{tab:a2}, we ablate state attributes and fusion to assess observation calibration. Pre-update correction error $E_{\rm corr}^{-}$ is 0.160 across ScanNet v2 variants. Removing provenance weakens correction, removing confidence increases historical drift, and removing uncertainty reduces correction gain. Unconditional fusion increases retention error most.

\noindent
\textbf{Impact of Incremental World Diffusion Generation.}
In Table~\ref{tab:a3}, we vary masking, hard freezing, and decoding on sampled ScanNet v2 scenes to evaluate incremental generation. Visibility-only masking corrects least. Global diffusion slightly improves LPIPS but increases historical drift and cost. Removing hard freezing increases drift. Full decoding matches quality and retention at higher cost.

\subsection{Discussion}
\label{sec:discussion}
\noindent
\textbf{Correction and Retention under New Evidence.}
In Table~\ref{tab:d1}, we compare update strategies on ScanNet v2 and Replica for correction and retention. Initial states share $E_{\rm corr}^{-}=0.160$ and 0.120, respectively. Correction targets valid, incorrect predictions. Extend-only retains history but leaves conflicts. Batch-Rebuild corrects more but increases drift. World-$\delta$ balances both by revision.

\noindent
\textbf{Selection under a Matched Update Budget.}
In Table~\ref{tab:d2_budget}, we evaluate update allocation on sampled ScanNet v2 scenes with 20\% of latent cells writable. Calibrated selection improves quality and correction over residual-only and random selection. Removing $U$ weakens correction. Visibility-only retains history slightly better but resolves fewer errors. Runtime depends on which chunks are updated. With the writable budget fixed, these results show the benefit of using calibrated evidence to prioritize corrections within the available update budget.



\noindent
\textbf{Dependency-Aware Incremental Decoding.}
In Fig.~\ref{fig7}, we compare decoder neighborhoods on ScanNet v2 with fixed latents and a fixed boundary region to assess agreement and efficiency. Boundary MAE measures depth error against reference geometry. Maximum RGB discrepancy measures deviation from full decoding. The default matches full decoding's reported MAE with negligible RGB discrepancy and lower runtime and memory use. Enlarging the halo and owner set increases cost without lowering MAE. Zero halo increases depth error and RGB discrepancy. These results support selecting owners by network dependencies and reusing unaffected chunks from the cache. Read-only halo cells supply context without enlarging the writable mask.

\noindent
\textbf{Long-Range Scene Expansion.}
Fig.~\ref{fig8} presents exploration with World-$\delta$ in two large indoor scenes through successive observations. When a previously unseen room is observed, our method automatically revises and expands the scene at room scale, generating plausible spatial transitions while preserving established content. As exploration proceeds, newly revealed regions are progressively integrated into a coherent global environment. These results demonstrate that World-$\delta$ supports both observation-driven correction and expansion within local scenes and coherent scene construction over larger spatial extents.

\noindent
\textbf{Limitations.}
World-$\delta$ assumes a largely static scene, so object motion can be confused with errors in earlier completions. Fusing observations from different motion states may leave stale geometry and inconsistent appearance. Future work could explore motion-aware masks to separate dynamic regions from background updates and object tracking to maintain evolving object states.

\section{Conclusion}
\label{sec:conclusion}
We present World-$\delta$, an observation-driven framework for continually expanding and correcting generated 3D worlds. Distinguishing observed evidence from provisional completions keeps uncertain content revisable while preserving reliable history. We encode geometric and semantic evidence in a persistent state tracking provenance, confidence, and generative uncertainty. Calibrated state deltas then guide masked latent diffusion and dependency-aware decoding, restricting generation and explicit assembly to affected regions. Experiments on single-image generation and continual updates show improved rendering quality and a favorable balance of observation-driven correction, historical retention, and computational cost.

{\small
\bibliographystyle{ieeenat_fullname}
\bibliography{main}
}
\end{document}

%% file: preamble.tex
\input{math_commands}

\usepackage{mathrsfs,xurl,multirow,float,placeins}
\usepackage{verbatimbox,diagbox,enumerate,threeparttable}
\usepackage{enumitem,array,setspace,makecell,wrapfig}
\usepackage[accsupp]{axessibility}

\definecolor{cvprblue}{rgb}{0.21,0.49,0.74}
\usepackage[pagebackref,breaklinks,colorlinks,allcolors=cvprblue]{hyperref}

\newcommand{\SchedulePageFloats}[2]{%
  \AddToHook{para/begin}{%
    \ifnum\value{page}=#1\relax
      \ifcsname WorldDeltaFloatsDone#1\endcsname\else
        \expandafter\gdef\csname WorldDeltaFloatsDone#1\endcsname{1}%
        #2%
      \fi
    \fi
  }%
}

%% file: math_commands.tex
\usepackage{amsmath,amsfonts,bm}

\def\eqref#1{equation~\ref{#1}}

\def\1{\bm{1}}

\DeclareMathAlphabet{\mathsfit}{\encodingdefault}{\sfdefault}{m}{sl}
\SetMathAlphabet{\mathsfit}{bold}{\encodingdefault}{\sfdefault}{bx}{n}

